\def\year{2022}\relax
\documentclass[letterpaper]{article} 
\usepackage{aaai22}  
\usepackage{times}  
\usepackage{helvet}  
\usepackage{courier}  
\usepackage[hyphens]{url}  
\usepackage{graphicx} 
\usepackage{natbib}  
\usepackage{caption} 
\DeclareCaptionStyle{ruled}{labelfont=normalfont,labelsep=colon,strut=off} 
\usepackage{amsmath}
\usepackage{amssymb}
\usepackage{booktabs,caption}
\usepackage[flushleft]{threeparttable}
\usepackage{color}
\title{Spiral Language Modeling}
\author {
    Yong Cao,
    Yukun Feng,
    Shaohui Kuang,
    Gu Xu \\
}

\affiliations{ByteDance \\ 
\{\centering 
yongc, yukunfeng, kuangshaohui, guxu\}@bytedance.com}

\begin{document}

\maketitle

\begin{abstract}
In almost all text generation applications, word sequences are constructed in a left-to-right (L2R) or right-to-left (R2L) manner, as natural language sentences are written either L2R or R2L. 
However, we find that the natural language written order is not essential for text generation.
In this paper, we propose Spiral Language Modeling (SLM), a general approach that enables one to construct natural language sentences beyond the L2R and R2L order. 
SLM allows one to form natural language text by starting from an arbitrary token inside the result text and expanding the rest tokens around the selected ones.
It makes the decoding order a new optimization objective besides the language model perplexity, which further improves the diversity and quality of the generated text.
Furthermore, SLM makes it possible to manipulate the text construction process by selecting a proper starting token.
SLM also introduces generation orderings as additional regularization to improve model robustness in low-resource scenarios.
Experiments on 8 widely studied Neural Machine Translation (NMT) tasks show that SLM is constantly effective with up to 4.7 BLEU increase comparing to the conventional L2R decoding approach.
\end{abstract}

\section{Introduction}
Natural language sentences are formed with tokens written either L2R or R2L.  
As a result, natural language sentences are represented by token sequences in their written order in most NLP applications.
In conventional language modeling, the joint probability of a token sequence $P(\textbf{x})$ is usually factorized along the time axis $P(\textbf{x})=\prod_{t}P(x_t|x_{<t})$.
The factorization implies a default ordering of L2R.
Accordingly, when applying the corresponding language model to text generation tasks, the generation ordering is naturally L2R.
However, constructing a natural language sentence is not necessarily an L2R approach in people's minds.
The construction process is more like starting from a critical cognitive element and filling in auxiliary words around it until a complete sentence with correct grammar.
Researchers have already conducted empirical studies on the sequence ordering for density estimation, and auto-regressive pre-training tasks \cite{uria2014deep,germain2015made,uria2016neural,yang2020xlnet}.

Permutation language modeling \cite{yang2020xlnet} is carried out to model sequence orderings in auto-regressive pre-training.
However, it may not be directly applied to text generation.
In-text generation, the result sequence length is unknown until a special end token is generated.
So a generation model may not foresee a token in the far end as we don't know whether the generation process will stop at the very next position.
Both L2R and R2L ordering holds the above property so that they are widely used in different text generation tasks.
Permutation language modeling does not hold the property as it allows to predict tokens in the far end.

\begin{figure}[t]
\centering
\setlength{\belowcaptionskip}{-0.9cm} 
\begin{center}
    \includegraphics[trim=0 0 0 0,clip,width=0.8\columnwidth]{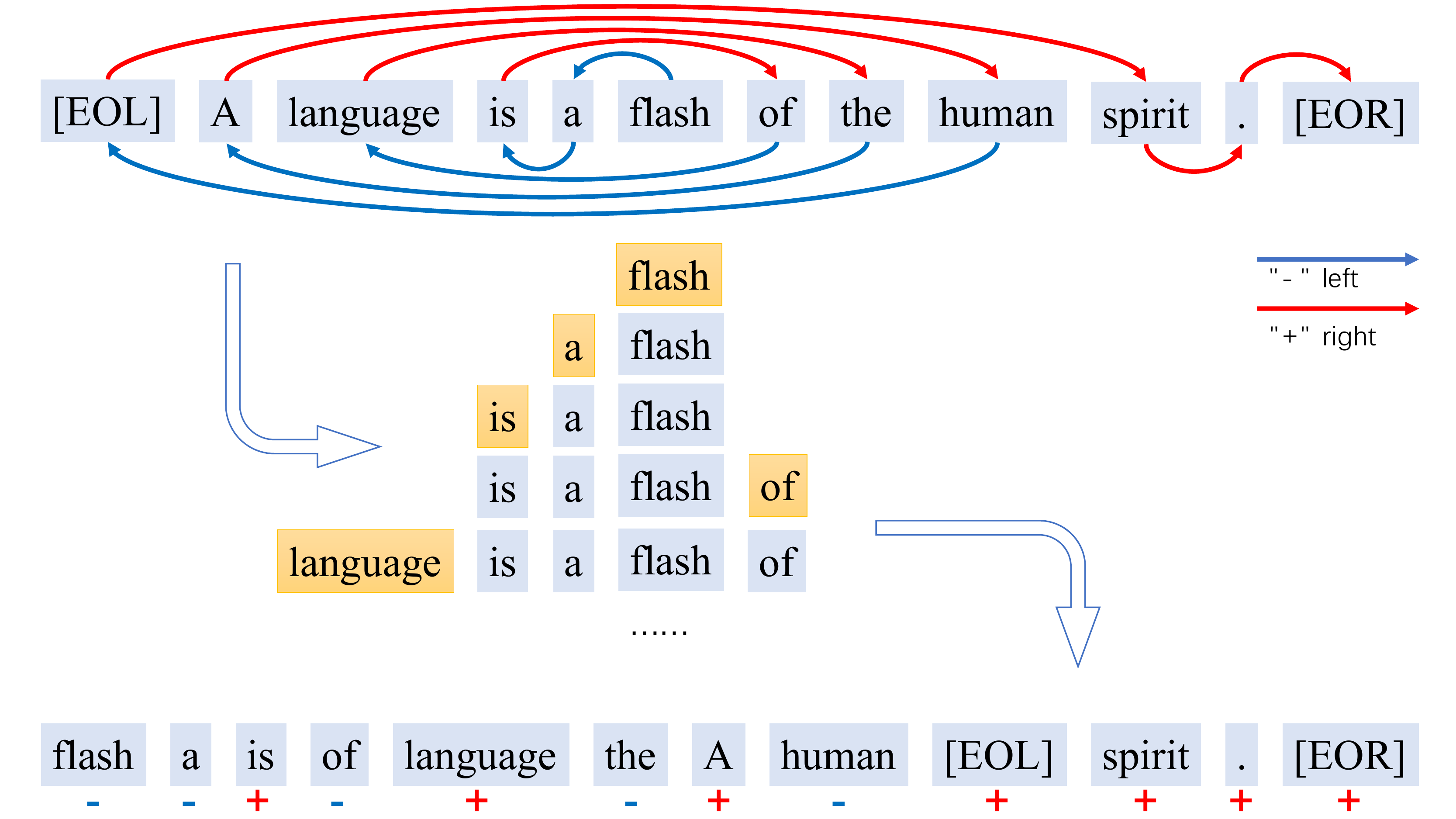}
\end{center}
\caption{An example of generation ordering in SLM}.
\label{slm_example}
\end{figure}

In this paper, we propose a general approach called Spiral Language Modeling that allows one to construct text sentences with any valid generation ordering.
Figure \ref{slm_example} shows an example of generation ordering.
The generation process starts from an arbitrary token inside the result sentence.
The selected token forms an initial segment.
The initial segment in the example is ``flash''.
The rest generation process constructs around the segment ``flash'' by arranging tokens on its right or its left following the arrows.
The high-lighted tokens in the middle of the figure are the appended ones at each step.
``a'' is appended to the left, then ``is'' is also appended to the left, while ``of'' is appended to the right, and so on so force.
We introduce two special tokens, ``[EOL]'' and ``[EOR]'', which stand for the left end and the right end, respectively.
The generation process stops when both end tokens are generated.
The sequence at the bottom of the figure is the actual generation ordering.
Specifically, if one starts from ``[EOL]'' and constantly appends tokens to its right until ``[EOR]'', the generation ordering is exactly L2R.
That is, L2R is just a special case of SLM generation ordering.
Similarly, we may get R2L ordering by starting from ``[EOR]''.
When we draw the construction arrows as shown in Figure \ref{slm_example}, these arrows form a set of spirals over the text.
This is why we call our approach Spiral Language Modeling.

When applying to text generation, SLM provides a family of valid generation orderings for one single text sentence.
In training, the SLM objective is the sequence probability expectation over all generation orderings.
During inference, different generation ordering may lead to the different result text.
Therefore, SLM actually provides a family of decoding orderings so that the text generation model may pick an optimal one from them.
As SLM may start from any token inside the result sentence, one can easily manipulate the decoding process to ensure the occurrence of any token or phrase.
The only thing one needs to do is to fix the token or phrase as the initial segment and let SLM infer the rest of the sentence.
Comparing to the conventional text generation approaches, SLM has the following distinctive advantages.
\begin{itemize}
    \item SLM provides an additional ordering dimension for the language models to optimize.
    \item SLM introduces generation orderings as additional regularization to mitigate over-fitting, as it prevents the language models from memorizing training data.
    \item SLM allows one to easily manipulate the decoding process.
\end{itemize}

We conduct empirical studies on Machine Translation (MT) tasks where the target language sentence decoding is a typical text generation problem.
Most of the conventional MT approaches follow the L2R decoding ordering.
We employ the same encoder and decoder network structure when comparing SLM to L2R ordering baselines.
The empirical study shows that SLM is constantly effective on 2 IWSLT'14 tasks and 6 IWSLT'16 tasks with an average 2.6 BLEU increase.
Specifically, SLM achieves significant improvement on IWSLT'16 CS$\to$EN and EN$\to$CS with 4.7 and 4.2 BLEU increases, respectively.
Here we summarize our contributions as follows.
\begin{itemize}
    \item We propose a novel language modeling approach, called Spiral Language Modeling.
    \item SLM has distinct advantages for text generation, including generation ordering optimization, training data regularization, and decoding process manipulation.
    \item We apply SLM to MT tasks and conduct detailed experiments to show its effectiveness.
\end{itemize}

\section{Related Work}
Most of the Neural Machine Translation (NMT) approaches decode the target language sentences through an L2R order \cite{vaswani2017attention,zhang2018asynchronous,zhou2019synchronous,Wang2020Encoding}.
Researchers are also interested in the research topics on sequence ordering and text decoding manipulation.

\textbf{Sequence Ordering} Sequence ordering is studied in the density estimation tasks ever since \citet{uria2014deep} propose to optimize the mean auto-regressive cost over all orderings of a given variable sequence.
\citet{germain2015made} further implement their order-agnostic training by permuting the input variable sequence with observed variables all before the masked ones.
\citet{uria2016neural} treat the orderings of the sequence variables as a stochastic variable
with a uniform distribution and optimize the expected likelihood over the ordering of variables.
Introducing ordering generally improves the density estimation effectiveness.
Recently, permutation language modeling is borrowed in the large-scale unsupervised auto-regressive model pre-training task XLNet in \citet{yang2020xlnet}.
The XLNet design a partial prediction task where normally the last 1/6 of the permuted sequence is masked for reconstruction.
To the best of our knowledge, there is no existing work discussing sequence ordering in text generation tasks.
\citet{bao2020unilmv2} also employed a pseudo-mask method to support different factorization orders in the Partially AutoRegressive model, in which the randomly sampled factorization orders are similar to permutation-based language modeling used by XLNet.

\textbf{Decoding Manipulation} Decoding text during generation naturally follows the text written order. However, employing additional decoding order generally introduces text generation diversity which leads to better performance.
\citet{finch2009bidirectional} use an R2L trained language model together with an L2R trained one and achieve BLEU gains constantly on all MT tasks.
\citet{zhang2013beyond} expand the bidirectional decoding idea to a mixture of the L2R and R2L ordering with a directed graphical model factorization.
NMT approaches such as \cite{liu2016agreement,sennrich2016edinburgh,hoang2017towards,sennrich2017university} run beam search for L2R and R2L models independently to get two n-best lists and re-score the union of the two lists to obtain the best one.
They all train separate forward and reverse decoders and require multiple rounds of decoding during inference.
Comparing to the above bi-directional decoding approaches, SLM is much neater as it explores all possible decoding directions that could maximize the result sequence scores in a single round of beam search.

Besides re-scoring, researchers are also trying to find a unified bi-directional searching approach.
\citet{zhang2018asynchronous} employ the reverse decoder as an auxiliary module for the forward decoder to foresee a part of the target text.
\citet{zhou2019synchronous} propose a synchronous bidirectional neural machine translation (SB-NMT) approach that generates an L2R sequence and an R2L sequence simultaneously and interactively.
\citet{zhousequence} carry out a decoding strategy that starts from both ends of a target sequence and generates two subsequences separately.
Moreover, \citet{yang2021smart} proposes a reversed strategy of the above one, which predicts a starting token first and generates the two subsequences on both sides separately.
All the above approaches simply try to find a better way to combine an L2R and an R2L language model.
Besides MT tasks, \cite{mou2015backward,mou2016sequence,li2018syntactically,liu2019bfgan} also employ bi-directional language models with well chosen decoding strategy in dialog generation tasks.
Our proposed SLM breaks the constraints that language modeling may only follow the L2R or R2L order.
Therefore, SLM provides a wide number of decoding strategies that the result decoding model can choose from.


\section{Approach}
SLM is a general approach for text generation. 
In this section, we first present the definition of SLM and the underlying theory.
We further introduce the SLM applications for the Neural Machine Translation (NMT) tasks.

\begin{figure*}[!ht]
\centering
\setlength{\belowcaptionskip}{-0.6cm} 
\begin{center}
    \includegraphics[trim=0 320 100 0,clip,width=2.0\columnwidth]{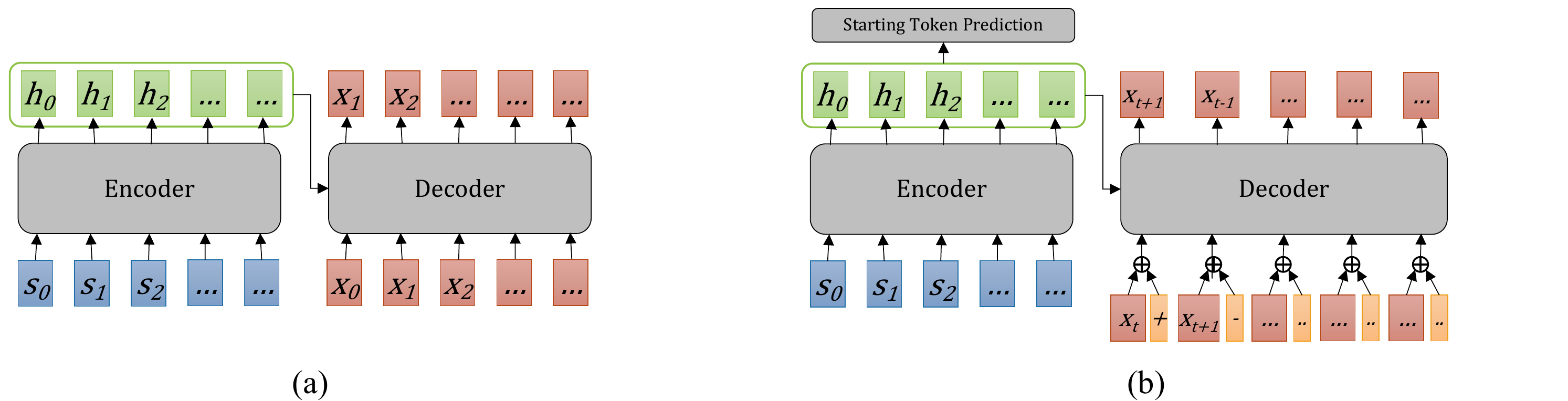}
\end{center}
\caption{(a) Conventional encoder-decoder network, (b) SLM encoder-decoder network}.
\label{slm_architecture}
\end{figure*}

\subsection{Spiral Language Modeling}
SLM is carried out specifically for text generation scenarios.
We borrow the permutation language modeling from \citet{yang2020xlnet} and change it to fit text generation applications.

\subsubsection{Language Modeling}
Consider a sequence of random variables, $X_1, X_2, ..., X_T$, each of which can take any value in a finite set $\mathcal{V}$.
The Language Modeling goal is to estimate the probability of any sequence $\textbf{x}=[x_1,x_2,...,x_T]$, where $T\geq1$ and $x_t \in \mathcal{V}$ for $t=1 ... T$, that is, to estimate the joint probability
\begin{equation}
    P(\textbf{x}=x_1,x_2,...,x_T)
\end{equation}
The above joint probability can be factorized along the time axis according to the law of total probability
\begin{equation}\label{p_factorization} 
P(\textbf{x})\&=\prod_{t}P(x_t|x_{<t})
\end{equation}
Language models introduce a set of parameters $\theta$ to approximate the conditional probability $p(x_t|x_{<t},\theta)$ by maximizing $p(\textbf{x})=\prod_{t}p(x_t|x_{<t},\theta)$.

Equation \ref{p_factorization} implies a default language construction order which is from left to right.
Let $\textbf{z}$ denote the ordering vector which depicts the construction order of a given language sentence.
The default ordering vector in Equation \ref{p_factorization} is $\textbf{z}=[1,2,...,T]$.
The factorization of the joint probability $P(\textbf{x})$ explicitly depends on the ordering $\textbf{z}$ as follows.
\begin{equation}\label{plm_factorization}
    P(\textbf{x})=\prod_{t}P(x_{z_t}|x_{z_{<t}})
\end{equation}
The ordering $\textbf{z}$ ranges from a full set of the permutation ordering $\mathbb{Z}_T$.
However, not all ordering is suitable for text generation.

\subsubsection{Generation Ordering}
When applying language models for text generation, the sequence length $T$ is unknown.
In the L2R and R2L generation approaches, the sequence length is predicted by introducing an explicit end token, such as ``[EOS]''.
The generation ordering $\textbf{z}$ should hold the property that the selection of $z_{t}$ given $z_{<t}$ must be valid for any possible length $T$.
For example, the ordering $z_{<t}=[1,2,3], z_{t}=5$ will result in an invalid factorization when the sequence length is actually 4.
That is, we cannot foresee a token in the far end during sequence generation as we don't know whether the generation process will stop at the next position.
Both L2R and R2L ordering holds the above property.
The text decoding alternatives mentioned in the related work section all follow the orderings that hold the above property.

SLM introduces a set of permutations that hold the above property. Specifically, a sequence is re-ordered by firstly picking an arbitrary token or a sequence of continue tokens inside the original one and keeps expanding the existing segment by appending the next left or right token until the segment meets the end tokens for both the left and the right direction. We introduce two special tokens, ``[EOL]'' and ``[EOR]'', representing the end-of-left and the end-of-right, respectively, when both tokens are decoded in $x_{z<t}$, the generation process stops.
Figure \ref{slm_example} shows an example of how a text sequence is generated in a valid generation ordering in SLM.
In fact, L2R and R2L orderings are two special cases of SLM.
L2R starts from the very left token and keeps appending the next right token while R2L does in a reversed manner.
Furthermore, the alternatives, which combine L2R and R2L language models in different ways, are all special cases of SLM.

\subsubsection{Objective}
Instead of maximizing a single joint probability, the objective of SLM is to maximize the expectation over all generation orderings.
\begin{equation}\label{slm_objective}
\begin{aligned}
    \max_\theta \mathbb{\textbf{E}}_{\textbf{z} \in \mathbb{Z}_{g}} \left [ \prod_{t}p_{\theta}(x_{z_t}|x_{z_{<t}}) \right ]
\end{aligned}
\end{equation}
where $\mathbb{Z}_{g}$ is the set of all generation orderings.
For a given text sentence $\textbf{x}$ with $T$ tokens, an estimated upper bound of $|\mathbb{Z}_{g}|$ is $T\cdot2^{T-2}$.
Therefore, it's intractable to go through $\mathbb{Z}_{g}$ to estimate equation (\ref{slm_objective}).
A straightforward approach is to apply a random sampling strategy.

A certain generation ordering actually defines a decoding policy which leads to an individual joint probability factorization over the learned parameters.
During inference, the ordering $\textbf{z}$ and the next token $x_{z_t}$ are evaluated together with a beam search approach to maximize $p_{\theta}(\textbf{x})$.
A proper decoding policy corresponding to the ordering $\textbf{z}$ will result in a better perplexity of the result sequence.
Therefore, we could employ proper sampling strategies during training to reinforce the selection of generation order.
We apply a two-stage sampling strategy in the NMT applications.
We first apply a random sampling strategy during the first half of the training steps.
We then sample the generation orderings that result in higher perplexities for the rest training steps.
The sampling strategies are further discussed in the empirical study session.

\subsection{Machine Translation}
NMT has attracted much attention ever since the seq-to-seq architecture \cite{sutskever2014sequence} is carried out.
Most of the popular NMT solutions employ encoder-decoder network architectures with different variations. 
The translation objective is the target language sentence probability conditioned on the source language sentence.
\begin{equation}\label{nmt_objective}
    \max_\theta \prod_{t}p_{\theta}(x_{t}|x_{<t}, \textbf{s}) 
\end{equation}
where $\textbf{s}$ is the source language sentence.
Machine translation with encoder-decoder architecture is a special text generation scenario where we can naturally apply SLM.
By introducing the generation ordering, equation \ref{nmt_objective} is changed to 
\begin{equation}\label{nmt_slm_objective}
\begin{aligned}
    \max_\theta \mathbb{\textbf{E}}_{\textbf{z} \in \mathbb{Z}_{g}} \left [ \prod_{t}p_{\theta}(x_{z_t}|x_{z_{<t}}, \textbf{s}) \right ]
\end{aligned}
\end{equation}
which is to maximize the expectation over all generation orderings of the target sentence given the source sentence.

\subsubsection{SLM Architecture}
Almost all decoder networks are implemented to follow a L2R decoding procedure.
RNN-based decoders recursively append RNN cells to the right of the existing RNN chain to generate the next token, while the transformer-based \cite{vaswani2017attention} decoders employ a special attention mask to prevent positions from attending to subsequent positions on their right.
Manipulating the text generation order is done by changing the sequences that fed into the decoder networks.
For example, feeding in a reversed target language text will result in an R2L text generation order.

\begin{table*}[t]
\centering
\setlength{\belowcaptionskip}{-0.2cm} 
  \begin{threeparttable}
     \begin{tabular}{lllllllll}
        \toprule
        \textbf{Model} & 
        \small{\textbf{DE-EN'14}} & 
        \small{\textbf{EN-DE'14}} &
        \small{\textbf{CS-EN’16}} &
        \small{\textbf{EN-CS‘16}} &
        \small{\textbf{FR-EN’16}} &
        \small{\textbf{EN-FR’16}} & 
        \small{\textbf{DE-EN‘16}} &
        \small{\textbf{EN-DE’16}} \\
        \midrule
        L2R Baseline & 
        \multicolumn{1}{c}{34.56}   &
        \multicolumn{1}{c}{28.34}   &
        \multicolumn{1}{c}{23.58}   &
        \multicolumn{1}{c}{16.54}   &
        \multicolumn{1}{c}{37.29}   &
        \multicolumn{1}{c}{38.12}   &
        \multicolumn{1}{c}{31.58}   &
        \multicolumn{1}{c}{26.24}   \\
        SLM+Random & 
        \multicolumn{1}{c}{35.23}   &
        \multicolumn{1}{c}{29.24}   &
        \multicolumn{1}{c}{28.08}   &
        \multicolumn{1}{c}{20.68}   &
        \multicolumn{1}{c}{37.91}   &
        \multicolumn{1}{c}{40.12}   &
        \multicolumn{1}{c}{32.61}   &
        \multicolumn{1}{c}{28.79}   \\
        SLM+Sampling & 
        \multicolumn{1}{c}{\textbf{36.18}}   &
        \multicolumn{1}{c}{\textbf{29.52}}    &
        \multicolumn{1}{c}{\textbf{28.26}}   &
        \multicolumn{1}{c}{\textbf{20.73}}    &
        \multicolumn{1}{c}{\textbf{39.14}}    &
        \multicolumn{1}{c}{\textbf{40.33}}    &
        \multicolumn{1}{c}{\textbf{33.87}}    &
        \multicolumn{1}{c}{\textbf{28.83}}    
        \\
        &
        \multicolumn{1}{c}{\textbf{(+1.62)}}   &
        \multicolumn{1}{c}{\textbf{(+1.18)}}    &
        \multicolumn{1}{c}{\textbf{(+4.68)}}   &
        \multicolumn{1}{c}{\textbf{(+4.19)}}    &
        \multicolumn{1}{c}{\textbf{(+1.85)}}    &
        \multicolumn{1}{c}{\textbf{(+2.21)}}    &
        \multicolumn{1}{c}{\textbf{(+2.29)}}    &
        \multicolumn{1}{c}{\textbf{(+2.59)}}    \\
        \bottomrule
     \end{tabular}
     \caption{SLM results comparing to L2R baselines}
     \label{iwslt_results}
  \end{threeparttable}
\end{table*}

We also want to implement SLM without changing the existing decoding procedures.
Given a text sentence $\textbf{x}=[x_1,x_2,...,x_T]$ and a specific generation ordering $z=[t, t+1, t-1, ...]$, we may form a unique sequence $\hat{\textbf{x}}=[\langle x_t,+ \rangle,\langle x_{t+1},-\rangle,\langle x_{t-1},+\rangle...]$ which encodes both tokens and their ordering.
Each element of the sequence is a tuple of the current token and the next construction direction.
``+'' refers to append the next token to the right while ``-'' refers to append to the left. 
Decoding the original text sequence $\textbf{x}$ with a given ordering $z$ is then equivalent to decoding the reformed $\hat{\textbf{x}}$ from left to right.

Although each element of $\hat{\textbf{x}}$ is a tuple, we only need to predict the token part and treat the direction part simply as input.
Firstly, there is no ground truth for the direction part as there are no preferred generation orderings.
Secondly, the direction part is only an auxiliary input to determine which token (on the left or the right) to predict.
For instance, the sub sequence $[\langle x_t,- \rangle,\langle x_{t-1},-\rangle]$ defines a segment $[x_{t-1},x_t]$.
The last tuple of the sequence holds a ``-'' direction.
It means that the next decoded token will be appended to the left of the current sequence and the next ground-truth token is $x_{t-2}$.
If we change the sub sequence to $[\langle x_t,- \rangle,\langle x_{t-1},+\rangle]$, the segment that it defines is still $[x_{t-1},x_t]$.
However, the next decoded token will be appended to the right of the sequence, and the corresponding ground-truth token is $x_{t+1}$ then.
To incorporate the direction information as part of the input of the model, We introduce the direction embedding $E^{dir_t}$.
Then the distributed representation of a tuple in $\hat{\textbf{x}}$ is a summation of the token embedding $E^{x_t}$ and the direction embedding $E^{dir_t}$.
\begin{equation}
    E^{\hat{x}_t} = E^{x_t} + E^{dir_t}
\end{equation}

\subsubsection{Starting Token Prediction}
Figure \ref{slm_architecture}-b shows an illustration of the SLM architecture.
Besides the modification of the decoder input, there is an additional structure appended to the encoder.
Unlike the fixed L2R generation ordering, where the starting token is always the very left one, SLM may start decoding from any token in the result sequence.
We need an independent network to estimate the very first factor $p_{\theta}(x_{z_1}|s)$.

\begin{figure}[h]
\centering
\setlength{\belowcaptionskip}{-0.6cm} 
\begin{center}
    \includegraphics[trim=0 250 490 0,clip,width=1.4\columnwidth]{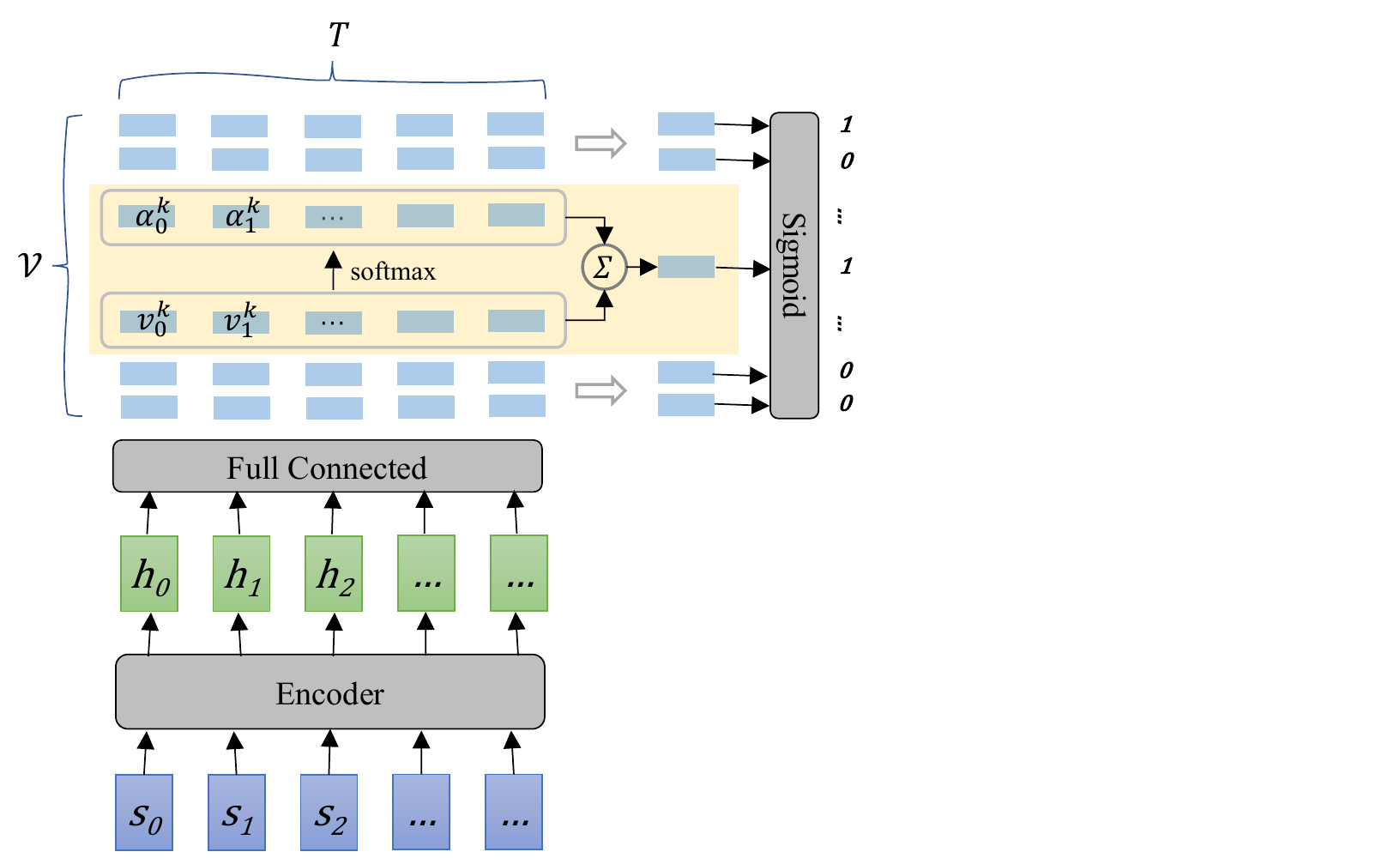}
\end{center}
\caption{Auto-alignment network}.
\label{autoalign}
\end{figure}

Estimating $p_{\theta}(x_{z_1}|s)$ is actually to determine which tokens will show up in the target language sentence given the source language sentence.
A straightforward solution is to build $|\mathcal{V}|$ binary classifiers for each is to estimate the possibility of the corresponding token occurring in the target sentence.
As the tokens inside the target sentence usually have a strong translation correlation to the corresponding tokens in the source sentence, we carry out an auto-alignment network to enhance the prediction performance.
Figure \ref{autoalign} shows the details of the starting token prediction network.

We put full connected networks right after the encoder output to compute the potentials over the target language vocabulary $\mathcal{V}$.
\begin{equation}
     v = \textbf{W}^{(H\times|\mathcal{V}|)}(\textbf{W}^{(H\times H)} h+\textbf{b}^{(H)})
\end{equation}
where $h$ is the encoder output and $v$ is the calculated potential.
The final potential of each vocabulary token $ \widetilde{v}^{(k)}$ is a weighted sum of $v$ along the time axis.
\begin{equation}\label{autoalign_attention}
    \alpha_{t}^{(k)}=\frac{\exp{v_{t}^{(k)}}}{\sum_{t} \exp{v_{t}^{(k)}}}
\end{equation}
\begin{equation}\label{autoalign_attention_sum}
    \widetilde{v}^{(k)} = \sum_{t} \alpha_{t}^{(k)} v_{t}^{(k)}
\end{equation}
where $v_{t}^{(k)}$ is the potential of the k-th token at time $t$, $\alpha_{t}^{(k)}$ is the normalized weight calculated from a softmax layer.
$\widetilde{v}^{(k)}$ is further fed to a Sigmoid layer where a binary classification cross entropy loss is applied afterward.
If the k-th vocabulary token occurs in the target sentence, it receives a label ``1'' and ``0'' otherwise.
The ultimate loss function is the summation of the starting token prediction loss and the translation loss.

\subsubsection{Inference}
We employ beam search during the translation inference. The probability of the initial token $p_{\theta}(x_{z_t}|s)$ is estimated with the starting token prediction network shown in Figure \ref{autoalign}.
One needs to try both directions when decoding the rest tokens.
Let's take the sentence in Figure \ref{slm_example} as an example.
Assuming that the beam size is 2, we now have top-2 candidates [``flash'', ``language''] after the first step of inference.
We need to try both left and right searching directions by appending direction input manually.
So we get 4 tuples, $[\langle flash,+ \rangle, \langle flash, - \rangle, \langle language, + \rangle, \langle language, - \rangle]$, as input for the next round of beam searching.

\section{Experiments}
We evaluate the effectiveness of SLM on the well-studied NMT tasks where the generation of the target language text is a typical text generation sub-task.
Comparing SLM to the conventional L2R decoding ordering, we find that SLM is consistently effective.

\begin{table}[h]
\centering
\setlength{\belowcaptionskip}{-0.4cm} 
\small
  \begin{threeparttable}
     \begin{tabular}{lllll}
        \toprule
        \textbf{Dataset} & 
        \textbf{Language} & 
        \textbf{Train} &
        \textbf{Valid} &
        \textbf{Test} \\
        \midrule
        IWSLT'14 &
        EN$\leftrightarrow$DE & 
        170 &
        7 &
        7 \\
        IWSLT'16 &
        EN$\leftrightarrow$CS & 
        114 &
        2 &
        2 \\
        IWSLT'16 &
        EN$\leftrightarrow$DE & 
        196 &
        2 &
        2 \\
        IWSLT'16 &
        EN$\leftrightarrow$FR & 
        220 &
        2 &
        2 \\
        \bottomrule
     \end{tabular}
     \caption{NMT datasets, numbers in thousand(k)}
     \label{nmt_datasets}
  \end{threeparttable}
\end{table}

\begin{figure*}
\centering
\setlength{\belowcaptionskip}{-0.8cm} 
\begin{center}
    \includegraphics[trim=0 0 0 0,clip,width=2.14\columnwidth]{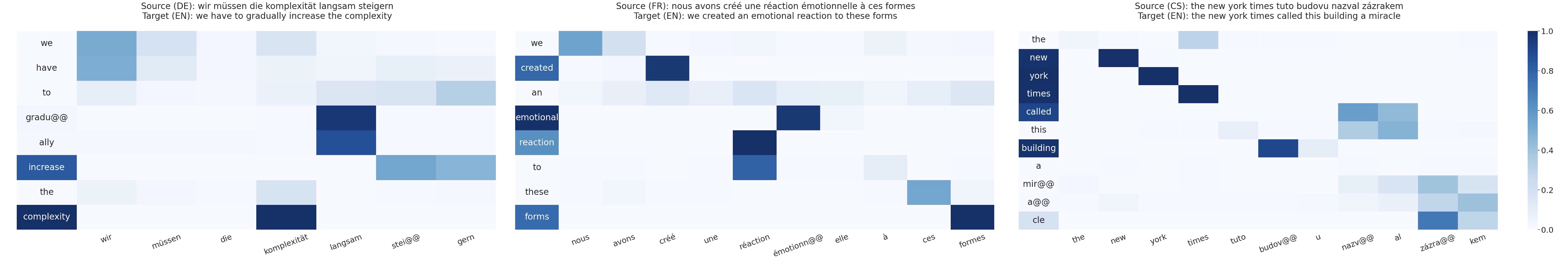}
\end{center}
\caption{Attention maps of start token prediction}.
\label{attentionmap_starttoken_3}
\end{figure*}

\subsection{Datasets and Settings}
The NMT datasets are listed in table \ref{nmt_datasets}, including IWSLT'14 English$\leftrightarrow$German (EN$\leftrightarrow$DE) and IWSLT'16 English$\leftrightarrow$\{Czech, French, German\}  (EN$\leftrightarrow$\{CS, FR, DE\}).
All data are pre-processed with the open-source tool Fairseq.
We employ Byte-Pair-Encoding (BPE) as the tokenization method with a vocab size of 10k for all datasets.

In the empirical studies, we focus on the comparison between SLM and the conventional L2R decoding baselines.
We choose the widely used Transformer \cite{vaswani2017attention} model with the transformer-base configurations for both SLM and L2R baselines.
We adopt the layer normalization and positional embedding as illustrated in \citet{raffel2020exploring}, in which the layer normalization in each block is set out of the residual path with the bias removed. 
In training, we use Adam optimizer with an initial learning rate of $9\times10^{-5}$.
The first 1000 update steps are warm-up steps.
The drop-out rate is set to 0.3.
During inference, we use a beam size of 5.
We introduce length penalty\cite{wu2016google} when calculating the scores of each beam search candidate.
We calculate the final BLEU score in a max order of 4 after removing BPE-tokens.
With the above settings, we reproduce very similar results on IWSLT'14 EN$\leftrightarrow$DE as reported in \cite{pham2021autodropout, wu2020sequence}.
The above settings are applied to both L2R baselines and SLM models, only that SLM introduces an additional starting token prediction part to the encoder as depicted in the last section.

\subsection{Result Analysis}
Table \ref{iwslt_results} shows the full experiment results on 8 translation tasks. 
L2R baselines and SLM reported results are adopted until full convergence or maximum update steps of 800k.
In the ``SLM+Random'' row, a uniform random sampling strategy is employed during the training stage.
In the ``SLM+Sampling'' row, we apply a two-stage sampling strategy.
During the first 90\% of the training process, we still employ the random sampling strategy, while in the rest 10\% update steps, we search for the best generation orderings based on the learned model. 
The detailed sampling strategy is discussed in the rest of the section.

As shown in the table, SLM significantly improves the NMT performance on all tasks.
We achieve an average increase of 2.6 BLEU scores.
Specifically, on the IWSLT'16 EN$\leftrightarrow$CS task, which has the smallest training dataset with only 114K sentence pairs, SLM outperforms the transformer baseline model with a 4.2 BLEU score.
We further study the properties of SLM from a different perspective in the rest of the section.

\subsubsection{Convergence}
\begin{figure}[h]
\centering
\setlength{\belowcaptionskip}{-0.5cm} 
\begin{center}
    \includegraphics[trim=0 0 0 0,clip,width=0.9\columnwidth]{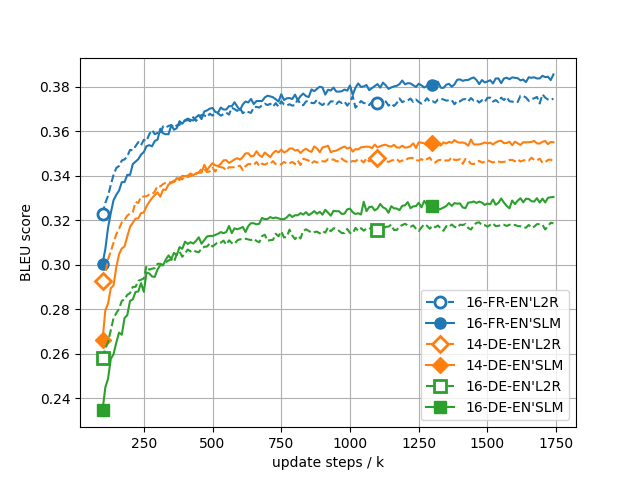}
\end{center}
\caption{Convergence curves}.
\label{convergence_curve}
\end{figure}

As we try to maximize the expectation of the joint probability over all valid generation orderings in SLM, the objective is more complex than the conventional L2R decoder objective.
For one single sentence pair, the target ground-truth text ordering is sampled on demand during training.
Therefore, the SLM model gets a different target ground truth every time it meets the same sentence pair. 
As a result, the SLM model converges slightly slower than the baseline L2R transformer model in the early training stage.
We compare the convergence of the baseline L2R transformer model and the SLM over 2 million update steps.
The experiments are conducted on the IWSLT'14 DE$\to$EN, IWSLT'16 (DE, FR)$\to$EN datasets, respectively.
The three convergence curves are shown in Figure \ref{convergence_curve}.
The baseline L2R models converge a little bit faster at the early training stage and stop growing at around 800k steps.
However, the SLM models keep growing better after outperforming the L2R baselines after around 400k steps.
The BLEU scores on the development set are still growing even when the update steps move close to 2 million.

\subsubsection{Low-Resource Scenario}
\begin{figure}[h]
\centering
\setlength{\belowcaptionskip}{-0.5cm} 
\begin{center}
    \includegraphics[trim=0 330 400 0,clip,width=1.3\columnwidth]{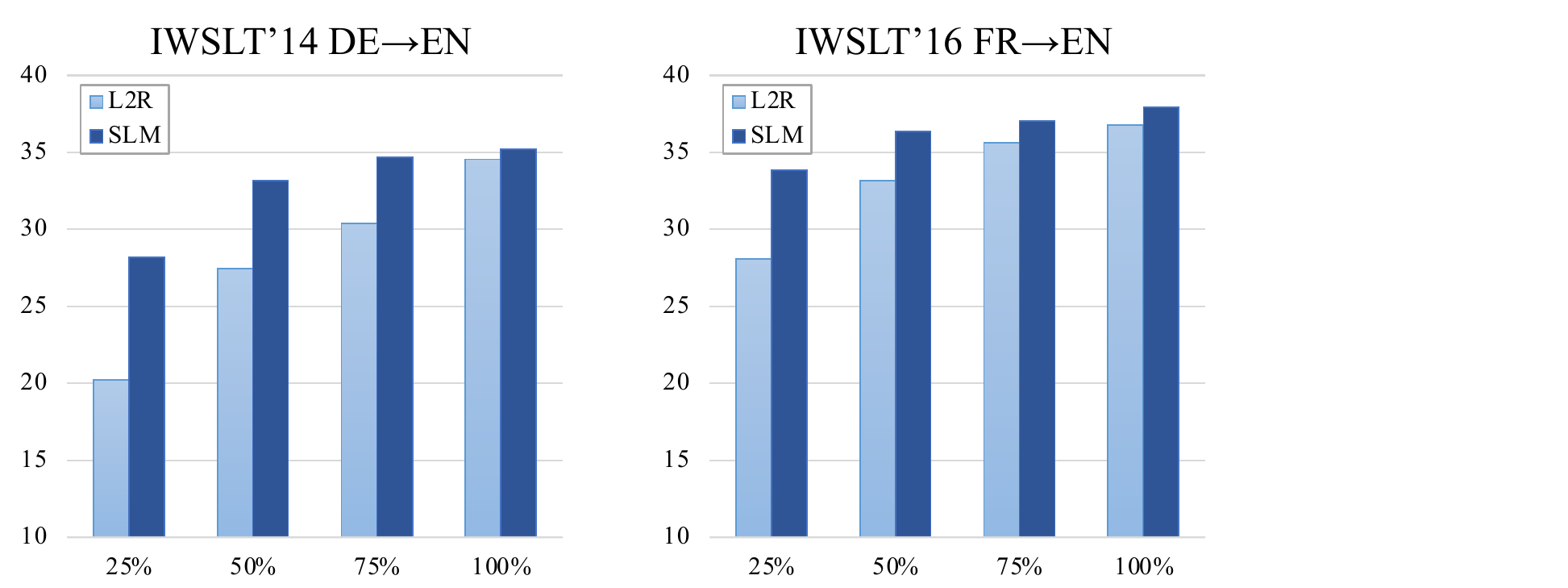}
\end{center}
\caption{Results on different training data portions}.
\label{low_resource_result}
\end{figure}

We notice that SLM achieves over 4 BLEU increases on the IWSLT'16 EN$\leftrightarrow$CS task, with the smallest training set with only 114K sentence pairs.
We further testify to the assumption that SLM may work better in low-resource scenarios.
We evaluate the conventional L2R model and SLM on IWSLT'14 DE$\to$EN and IWSLT'16 FR$\to$EN, with 1/4, 1/2, 3/4 of training data, respectively.
The results are shown in Figure \ref{low_resource_result}.
The performance gaps between the SLM models and the L2R baselines go bigger as there are less training data.
Specifically, the SLM model outperforms its L2R baseline on IWSLT'14 DE$\to$EN with an increase of close to 8 BLEU scores with only 1/4 of the full dataset (about 42K sentence pairs).
Comparing to the full set, SLM only has a 7 BLEU score decrease while the L2R baseline losses 15 BLEU scores with 1/4 training data.
We get similar observations on IWSLT'16 FR$\to$EN.
SLM introduces randomness to the training data by employing different generation orderings. SLM prevents the model from memorizing the training data and mitigates over-fitting problems, especially in low-resource scenarios. 

\begin{table*}[t]
\centering
\setlength{\belowcaptionskip}{-0.5cm} 
  \begin{threeparttable}
     \begin{tabular}{ll}
        \toprule
        \textbf{DE$\to$EN Example} &
        \\
        \midrule
        Source &
        \textit{wir werden fähig sein, unsere ideen direkt auf die digitalen medien abzuladen.} \\
        Ground-truth &
        \textit{we‘re going to be able to dump our ideas directly to digital media.} \\
        L2R baseline &
        \textit{we're going to be able to download our ideas directly to digital media.} \\
        SLM &
        \textit{we'll be able to transfer our ideas directly to the \textbf{digital} media.} \\
        \small {generation order} &
        \small {\textit{digital/+ media/- the/- to/- directly/- ideas/+ ./+ [EOR]/- our/- transfer/- to/- able/- be/- 'll/- we/- [EOL]}} \\
        SLM with fixed start &
        \textit{we'll be able to \textbf{dump} our ideas directly to digital media.} \\
        \small {generation order} &
        \small {\textit{dump/- to/+ our/- able/- be/+ ideas/+ directly/+ to/- 'll/- we/+ digital/+ media/+ ./+ [EOR]/- [EOL]}} \\       
        \bottomrule
     \end{tabular}
     \caption{Translation Examples from IWSLT'16 DE$\to$EN (the bold font tokens are the starting tokens)}
     \label{two_examples}
  \end{threeparttable}
\end{table*}

\subsubsection{Starting Token Prediction}
As SLM can generate result text starting from any token inside it, the model needs to decide which starting token may lead to a better result.
We employ an auto-alignment network structure as shown in Figure \ref{autoalign} to predict starting tokens. 
We directly predict which tokens will occur in the target language text given a source language sentence.
There is a tricky problem when predicting which tokens will occur in the target sentence.
The stop words usually have larger term frequencies which will result in a higher occurrence probability.
However, starting from a stop word does not provide much useful information.
So we exclude stop words when constructing ground-truth for the starting token prediction networks.
We filter stop words with the NLTK library.
The Sigmoid output value at each vocabulary index is an approximation of $p_{\theta}(x_{z_1}|\textbf{s})$, the probability of the corresponding token occurring in the target sentence.
The estimated probability is further used in the downstream decoding beam search as part of the joint probability factorization.

Starting token prediction is a multi-task learning approach.
The auto-alignment network is expected to generate a proper source sentence representation for reasoning whether a specific target token may be involved in the translated sentence.
The attention mechanism depicted in Equation \ref{autoalign_attention} and \ref{autoalign_attention_sum} is designed for this purpose.
The source sentence representation for different target tokens is different.
Figure \ref{attentionmap_starttoken_3} shows an example of the attention map.
We can see that the proposed attention mechanism can automatically align the translation token pairs among the translation sentence pairs.
For example, the English token ``complexity'' in the left attention map has a very big attention score at the German token ``komplexität''.

\subsubsection{Decoding Manipulation}
SLM allows the decoder to start from any token or any set of continuous tokens. The decoding process can be manipulated by selecting a proper starting token manually during the inference stage.
This feature allows one to preserve any token that s/he would like to put in the result sentence.
We put a DE$\to$EN example in Table \ref{two_examples} to show how we can manipulate the translation results.
The SLM model selects ``digital'' as the starting token after evaluating with a beam search strategy.
Suppose we've already had a candidate word, for example, "dump" to represent how the idea transferred to the media. In that case, we can manually start from "dump'' and get a result sentence much closer to the ground-truth sentence with the SLM model.

\subsubsection{Sampling Strategy}
SLM requires a sampling strategy during the training stage so that we may approximate the joint probability expectation.
A straightforward strategy is to sample generation orderings randomly.
However, each generation ordering defines a decoding policy, leading to an individual sequence joint probability factorization over the learned parameters.
Good decoding policies result in the better perplexity of the result sequence compared to the fair ones.
In the training stage, we generally want a sampling strategy to reinforce the selection of the better generation ordering that corresponds to better decoding policies.

\begin{table}[h]
\centering
\small
\setlength{\belowcaptionskip}{-0.4cm} 
  \begin{threeparttable}
     \begin{tabular}{lll}
        \toprule
        \textbf{Strategy} & 
        \textbf{Samples} & 
        \textbf{BLEU}  \\
        \midrule
        L2R &
        \multicolumn{1}{c}{1}   &
        \multicolumn{1}{c}{34.56}   \\
        Random & 
        \multicolumn{1}{c}{1}   &
        \multicolumn{1}{c}{35.23}   \\
        Sampling by fix top-1 starting token & 
        \multicolumn{1}{c}{1}   &
        \multicolumn{1}{c}{35.89}   \\
        Sampling by fix top-3 starting token & 
        \multicolumn{1}{c}{3}   &
        \multicolumn{1}{c}{\textbf{36.18}}   \\
        \bottomrule
     \end{tabular}
     \caption{Sampling strategies comparison}
     \label{sampling_experiment}
  \end{threeparttable}
\end{table}

We apply a two-stage sampling strategy.
In the first 90\% of the training process, we still employ the random sampling strategy.
In the rest 10\% of the training steps, we fix the top-k ranked start tokens returned by the starting token prediction networks. 
For each top-k token, we randomly select the rest orderings to speed up the sampling process. 
Table \ref{sampling_experiment} shows the comparison among different strategies on the IWSLT'14 DE$\to$EN task.
The best BLEU score is achieved by sampling over top-3 starting tokens.
Keeping more top-k tokens does not improve the final performance anymore.
When keep top-3 starting tokens, one sentence pair is replicated 3 times with different decoding ordering.
The rest training steps should at least go through all training data once.

We also summarize the reported start-of-the-art (SOTA) results on the IWSLT'14 DE$\to$EN task in table \ref{iwslt14_sota}.
We employ the two-stage sampling strategy to get the reported SLM results.
The SLM result in the last row is very much close to the result reported by \citet{wu2020sequence}. 
As the main purpose of this paper is to discuss how orderings affect text generation effectiveness in SLM, we do not put effort into integrating existing independent approaches to outperform the SOTA.
The below table provides a reference of how much SLM advances the L2R baseline on the IWSLT'14 DE$\to$EN task.

\begin{table}[h]
\centering
\setlength{\belowcaptionskip}{-0.5cm} 
\small
  \begin{threeparttable}
     \begin{tabular}{lll}
        \toprule
        \textbf{Model} & 
        \textbf{DE$\to$EN} \\
        \midrule
        Transformer Base\cite{vaswani2017attention} & 
        \multicolumn{1}{c}{34.56}   \\
        Adversarial training\cite{wang2019improving} &
        \multicolumn{1}{c}{35.2} \\
        Mixed Representations\cite{wu2020sequence} &
        \multicolumn{1}{c}{\textbf{36.4}} \\
        AutoDropout\cite{pham2021autodropout} &
        \multicolumn{1}{c}{35.8} \\
        Smart-Start Decoding\cite{yang2021smart} &
        \multicolumn{1}{c}{35.61} \\
        \midrule
        SLM+Sampling 800k-steps & 
        \multicolumn{1}{c}{36.18}   \\
        SLM+Sampling 2 million-steps & 
        \multicolumn{1}{c}{36.31}   \\
        \bottomrule
     \end{tabular}
     \caption{Comparison to SOTA on IWSLT'14 DE$\to$EN.}
     \label{iwslt14_sota}
  \end{threeparttable}
\end{table}




\section{Conclusion}
In this paper, we present a general approach called SLM that enables one to generate natural language sentences beyond the L2R and R2L orderings. 
The generation ordering employed by SLM constructs natural language text by starting from an arbitrary token inside the result text and then appending tokens around the constructed segment.
By introducing generation ordering, SLM includes a new optimization objective besides the language model perplexity, which further improves the diversity and quality of the generated text.
As SLM samples the generation order during the training stage, SLM introduces additional regularization to improve model robustness.
The experiments also show that SLM can work well under extreme low resource scenarios in the NMT tasks.
Furthermore, one may easily manipulate the text construction process by manually choosing the initial decoding order.
We conduct experiments on 8 widely studied NMT tasks.
The results show that SLM is constantly effective comparing to the conventional L2R decoding approach.

\bibliography{ref}

\end{document}